\documentclass[twoside,leqno,twocolumn]{article}

\usepackage{ltexpprt}
\usepackage{hyperref}

\usepackage{cite}
\usepackage{amsmath,amssymb,amsfonts}
\usepackage{algorithmic}
\usepackage{graphicx}
\usepackage{textcomp}
\usepackage{xcolor}
\usepackage{booktabs}
\usepackage{cleveref}
\usepackage{xcolor}

\begin{document}
\newtheorem{definition}{Definition}


%
\newcommand\relatedversion{}
\renewcommand\relatedversion{\thanks{This is the full version of the paper in SDM23}} 


\title{Eco-PiNN: A Physics-informed Neural Network for Eco-toll Estimation\relatedversion}

\author{Yan Li\thanks{Yan Li and Mingzhou Yang contributed equally to this paper.} \thanks{University of Minnesota. \{lixx4266, yang7492, eagon012, farha043,  wnorthro, shekhar\}@umn.edu}
\and Mingzhou Yang\footnotemark[2] \footnotemark[3]
\and Matthew Eagon\footnotemark[3]
\and Majid Farhadloo\footnotemark[3]
\and Yiqun Xie\thanks{University of Maryland. \{xie@umd.edu\}}
\and William F. Northrop\footnotemark[3]
\and Shashi Shekhar\footnotemark[3]}


\date{}

\maketitle


\fancyfoot[R]{\scriptsize{Copyright \textcopyright\ 2023 by SIAM\\
Unauthorized reproduction of this article is prohibited}}





\begin{abstract}
\small\baselineskip=9pt
The eco-toll estimation problem quantifies the expected environmental cost (e.g., energy consumption, exhaust emissions) for a vehicle to travel along a path. This problem is important for societal applications such as eco-routing, which aims to find paths with the lowest exhaust emissions or energy need. The challenges of this problem are three-fold: (1) the dependence of a vehicle's eco-toll on its physical parameters; (2) the lack of access to data with eco-toll information; and 
(3) the influence of contextual information (i.e. the connections of adjacent segments in the path) on the eco-toll of road segments.
Prior work on eco-toll estimation has mostly relied on pure data-driven approaches and has high estimation errors given the limited training data.
To address these limitations, we propose a novel Eco-toll estimation Physics-informed Neural Network framework  (Eco-PiNN) using three novel ideas, namely, 
(1) a physics-informed decoder that integrates the physical laws governing vehicle dynamics into the network,
(2) an attention-based contextual information encoder, and
(3) a physics-informed regularization to reduce overfitting.
Experiments on real-world heavy-duty truck data show that the proposed method can greatly improve the accuracy of eco-toll estimation compared with state-of-the-art methods.
\end{abstract}

\begin{keywords}
\textit{\textbf{Keywords:} eco-toll estimation, physics-informed machine learning, spatiotemporal data mining}
\end{keywords}

\section{Introduction}
\begin{figure}[t]
\centering
\includegraphics[width=8.5cm]{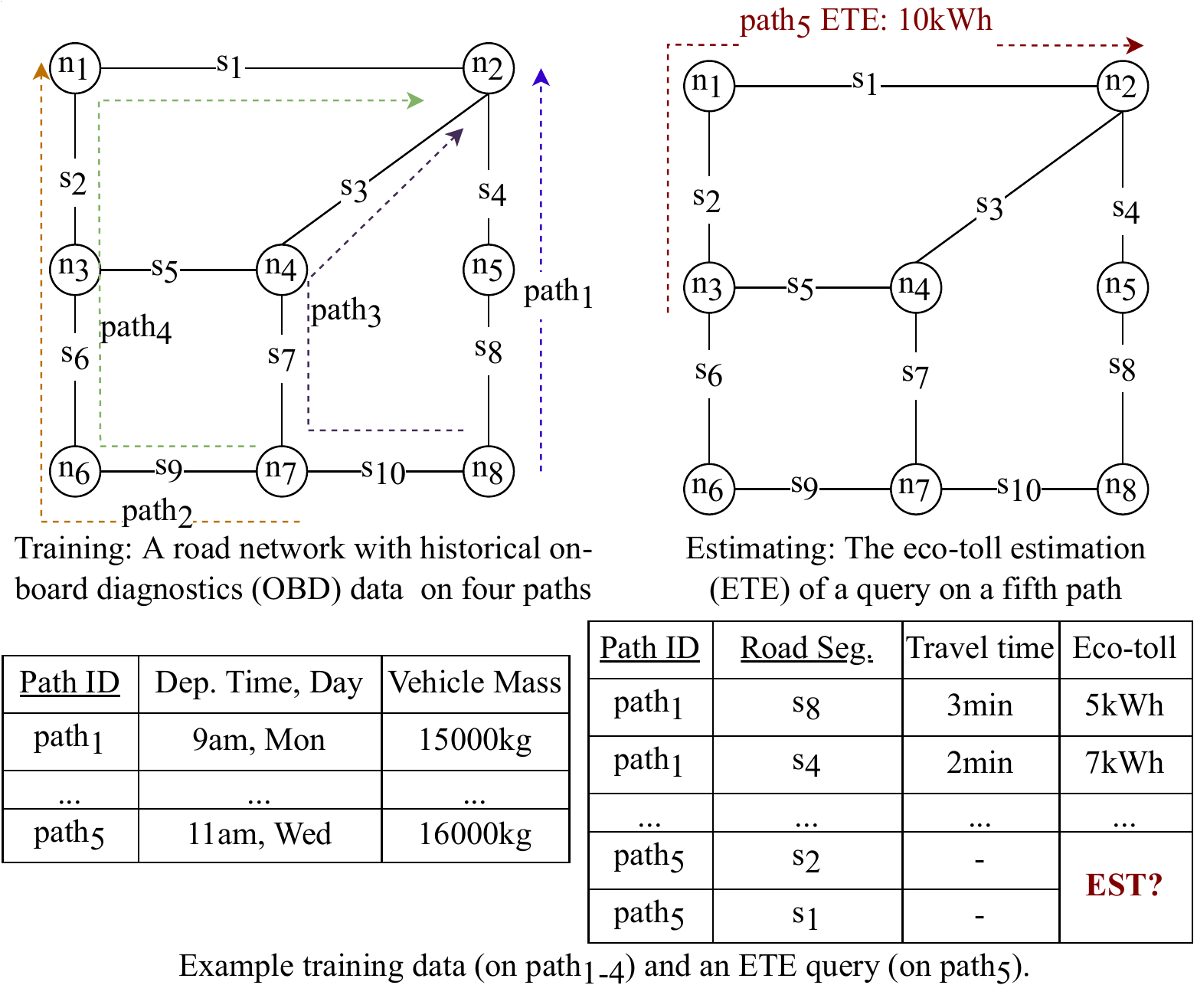}
\vspace*{-2.7\abovedisplayskip}
\caption{An example of the ETE problem with training vehicle OBD data attributes and an ETE query.}
\vspace*{-1.9\abovedisplayskip}
\label{fig:input}
\end{figure}

The development of on-board diagnostics (OBD) systems, which provide vehicle self-diagnosis and reporting capabilities, offers a transformative way to monitor the real-world functionality of vehicles.
Using historical OBD attributes as training data, the eco-toll estimation (ETE) problem aims to quantify the expected environmental cost (e.g., energy consumption, fuel consumption, exhaust emissions, etc.) for a vehicle given a query path and a user-specified departure time. Figure \ref{fig:input} shows an example of the ETE problem with historical OBD data on four paths ($\textrm{path}_{1-4}$) and one query composed of $\textrm{path}_5$, departure time, and the mass of a vehicle. This problem is of significant societal importance because it is an indispensable function of eco-routing, which aims to identify the most environmentally friendly travel route between two locations on a road network. Solving the ETE problem contributes to saving energy and mitigating transportation's impact on the environment and public health.





The challenges of this problem are three-fold. First, unlike common metrics of path selection such as distance and travel time, a vehicle's eco-toll is affected by the vehicle's physical parameters (e.g., vehicle weight, size, powertrain and power). 
Second, the paucity of available eco-toll data makes it challenging to develop accurate eco-toll estimation models. 
Most studies on eco-toll estimation models \cite{holden_trip_2018,huang_eco-routing_2018} are conducted on data generated from vehicle simulators. 
These simulators require second-by-second vehicle velocity profiles as a key input. Due to high cost, however, most large scaled mobile sensors have low sampling rates in practice, which greatly limits the availability of data for testing and training \cite{shan2018vehicle}.
Finally, the eco-toll on one road segment is influenced by the contextual information (i.e. the connections of adjacent segments) of the path. 
For example, a vehicle on a highway will incur an extra eco-toll for acceleration if it just enters from an entrance ramp. 


Most related work on eco-toll estimation is based on purely data-driven methods. For example, Huang and Peng proposed a Gaussian mixture regression model to predict energy consumption on individual road segments \cite{huang_eco-routing_2018}. The U.S. National Renewable Energy Laboratory (NREL) proposed a lookup-table-based method, which lists energy consumption rate by category of road segments \cite{holden_trip_2018}.  
However, travel eco-toll is influenced by many physical vehicle parameters (e.g. mass, shape, drive cycle, velocity profile, etc.). Thus, these purely data-driven methods have met with limited success due to their large eco-toll data requirements and have high estimation errors given the limited training
data.
To produce physically consistent results, Li et al. introduced a physics-guided K-means model \cite{li2018physics}, however it only provides results on paths with historical OBD data.

Much research has been conducted on other travel metrics (e.g., travel time). For example, Fang et al. proposed a contextual spatial-temporal graph attention network (ConSTGAT) \cite{fang2020constgat}, which contains a graph attention mechanism to extract the joint relations of spatial and temporal information and uses convolutions over local windows to encode contextual information of road segments. 
However, they do not consider the influence of a vehicle's physical parameters on its eco-toll, and also require large amounts of training data.
More details about the related work are in 
Appendix \ref{appendB}.

In this work, we propose an eco-toll estimation physics-informed neural network (Eco-PiNN) framework to address the ETE problem. Our main contributions are as follows:
(1) We propose a physics-informed decoder that integrates physical laws governing vehicle dynamics into Eco-PiNN.
(2) We propose an attention-based contextual information encoder to capture a path's contextual information.
(3) 
We introduce a physics-informed regularization (specifically, a jerk penalty) to guide the training of Eco-PiNN.  
(4) We conduct extensive experiments on real-world heavy-duty truck datasets, showing that Eco-PiNN outperforms the state-of-the-art models.

Purely data-driven machine learning (ML) models often suffer from limited success in scientific domains because of their large data requirements, and inability to produce physically consistent results \cite{Willard2022IntegratingSK}. Thus, research communities have begun to explore integrating scientific knowledge with ML in a synergistic manner. This kind of work is being pursued in diverse disciplines, such as climate science \cite{faghmous2014big},  biological sciences \cite{alber2019integrating}, etc. Our work represents the first effort to propose a model that leverages the physical laws of vehicle dynamics with neural networks to address the challenges in the ETE problem.
As summarized in Table \ref{tab:applicationdomain1}, the proposed model can also be generalized to estimation tasks on other
application areas where spatial graphs are defined, such as predicting electric power losses on transmission lines in electricity grids.


\begin{table}[]
\centering
\caption{Use cases of proposed model.}
\label{tab:applicationdomain1}
\begin{tabular}{@{}cc@{}}
\toprule
Application Area & Example Use Cases  \\ \midrule
Transportation & Eco-toll estimation
 \\ 
Electricity & Electricity grid loss estimation
\\ 
Environment & River flow estimation
\\
Computer Network & Internet traffic estimation
 \\ \bottomrule
\end{tabular}
\end{table}


In this paper, we only consider variables contained in existing OBD data. Other components which can influence the eco-toll but are either difficult to extract or are not typically found in OBD data (e.g., driver behavior, weather conditions, auxiliary power from HVAC system,, etc.) are not considered here. Computational complexity analysis is also outside the scope of this paper.

\section{Preliminaries}\label{preliminary}
\subsection{Notations and Definitions}

\begin{definition}
A \textbf{road network} refers to a weighted directed graph ($\mathcal{G} = (\mathcal{S}, \mathcal{N})$) modeling a road system in a study area, where $\mathcal{S}$ is a road segment set and $\mathcal{N}$ is a node set. Each $s_i \in \mathcal{S}$ represents a road segment (e.g., $s_1$ in Figure \ref{fig:input}), and a node $n_{i} \in \mathcal{N}$ represents a road intersection shared by segments (e.g., $n_1$ in Figure \ref{fig:input}). 
\end{definition}

\begin{definition}
A \textbf{path} is a sequence of road segments (e.g., in Figure \ref{fig:input}, $path_3 = [s_{10}, s_7, s_3]$). The $i$th segment of the path is denoted by $path(i)$ (e.g. $path_3(1) = s_{10}$). A \textbf{path length} is defined as the number of road segments in a path (e.g. $length(path_3)=3$).
A \textbf{subpath} is a path that makes up a larger path in the graph (e.g., $[s_{10}, s_7]$ is a subpath of $path_3$).

\end{definition}


\begin{definition}
An eco-toll estimation (ETE) \textbf{query} is represented by a three-element tuple $qry=(path_i, t_0, vp)$, where $path_i$ is the query path, $t_0$ is the departure time, and $vp$ is the vehicle's physical parameters (e.g., Figure \ref{fig:input} shows a query on $path_5$ of a 16-ton vehicle with departure time at 11am on Wednesday).
\end{definition}

\begin{definition}
\textbf{On-board diagnostics (OBD) Attributes}. Raw OBD data contain a collection of multi-attribute trajectories and the physical parameters of the corresponding vehicles. In this paper, each trajectory is map-matched to a path in the road network associated with a few OBD attributes to train the model, namely, the departure time, vehicle mass, travel eco-toll and travel time on each road segment (e.g., Figure \ref{fig:input} shows OBD training data on four paths).  

\end{definition}


\subsection{Problem Definition}

The eco-toll estimation problem is defined as follows:
\begin{itemize}
    \item \textbf{Input:} An ETE query $qry$ composed of a query path in a road network, a departure time $t_0$, and a vehicle's physical parameters $vp$.
    \item \textbf{Output:} The estimated eco-toll of the query $qry$.
    \item \textbf{Objective:} Minimize the estimation error.
    \item \textbf{Constraints:} In both training and testing datasets, the OBD data, including trajectories and physical parameters of vehicles,  are drawn from the same distribution.
\end{itemize}

In this work, we use energy consumption as a proxy for eco-toll. Other measures of eco-toll (e.g. fuel consumption, exhaustion emissions, etc.) can also be calculated from the energy consumption given the vehicle physical parameters (e.g. fuel type) \cite{davis2021transportation}. We only consider the variables within existing OBD data, so we assume there is no distribution shift between training and testing datasets.

\section{Proposed Approach}\label{approach}

\begin{figure}[t]
\includegraphics[width=7.9cm]{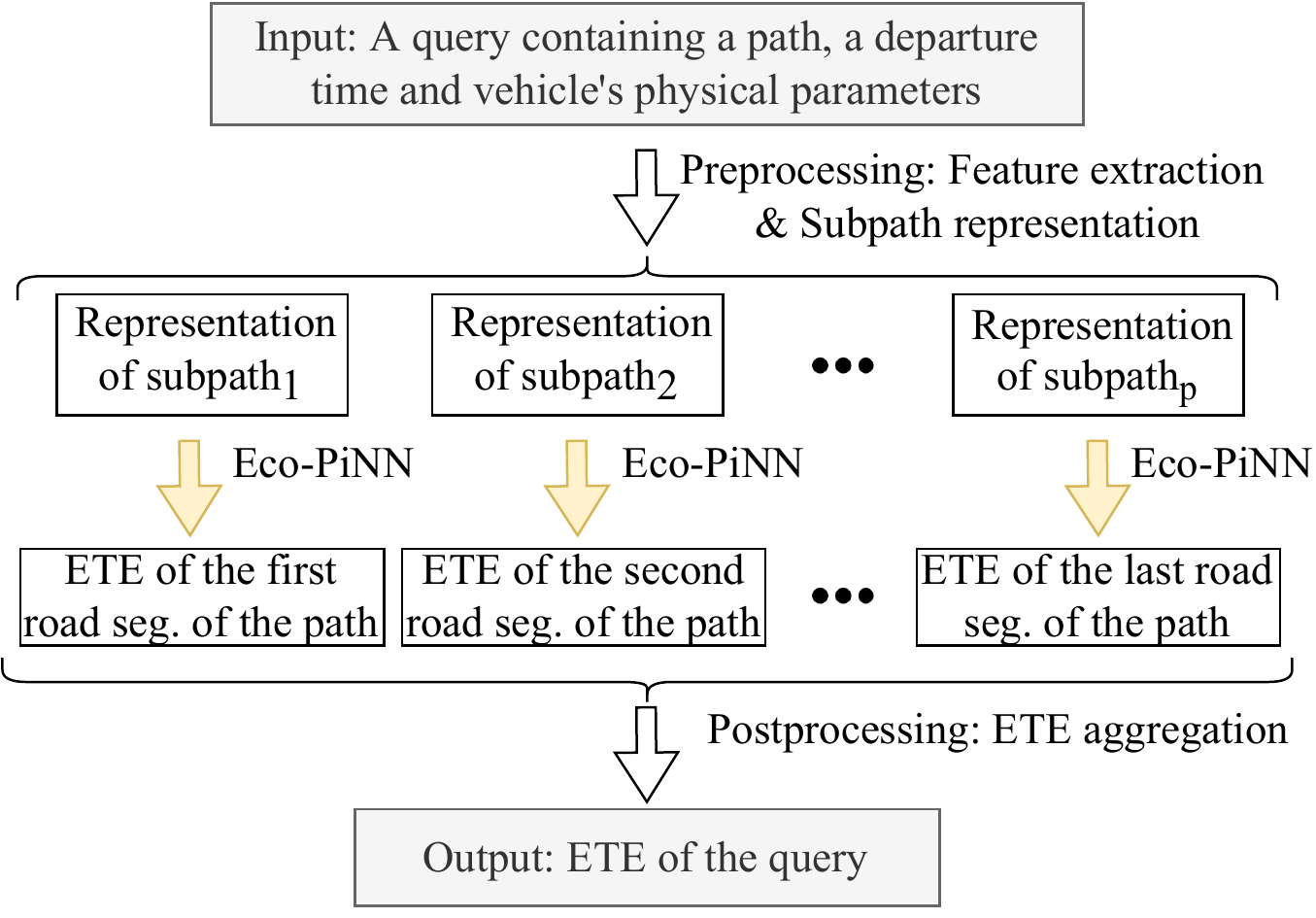}
\vspace*{-1\abovedisplayskip}
\caption{Overall framework for estimating a vehicle's eco-toll on a query path.}
\vspace*{-2.6\abovedisplayskip}
\label{fig:solFrame}
\end{figure}

Figure \ref{fig:solFrame} illustrates the overall framework architecture of the proposed solution. First, a \textbf{data preprocessing} module processes features extracted from an ETE query and a road network, and represents the query path as a sequence of subpaths. Then we propose a novel \textbf{Eco-PiNN} framework to estimate the eco-toll on a segment given the representation of the corresponding sub-path. Finally, in the \textbf{postprocessing} stage, the eco-toll estimation of segments of the path are aggregated together to generate the ETE of the query. In this paper, we use the sum operation to do the aggregation.



\subsection{Preprocessing}\label{sec:preprocessing}
The preprocessing module extracts and aggregates three different types of features from the ETE query.


\begin{figure}[t]
\includegraphics[width=8.5cm]{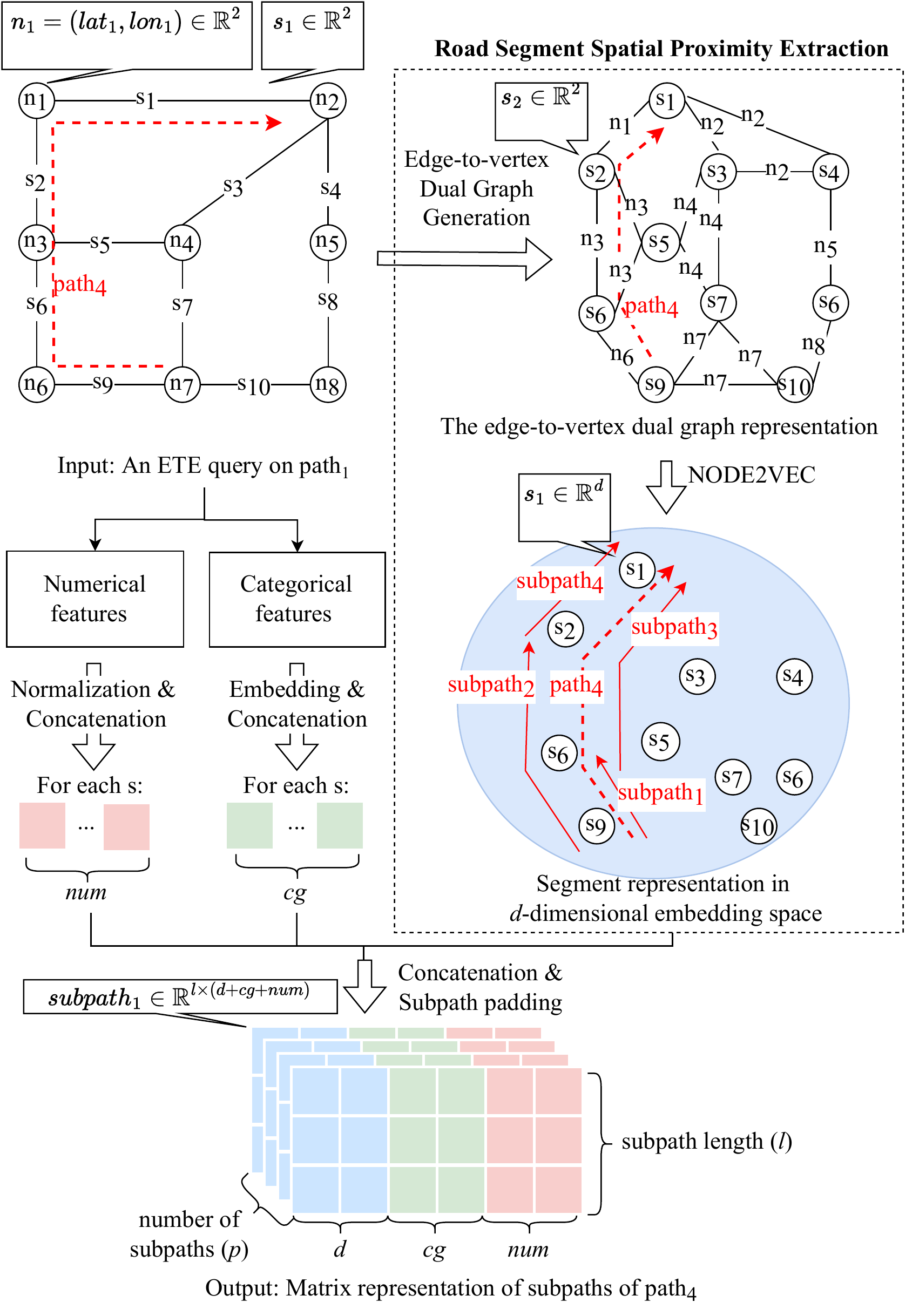}
\vspace*{-2.7\abovedisplayskip}
\caption{Preprocessing with road segment spatial proximity extraction and subpath representation using $path_4$ as an example (subpath length $l=3$).}
\vspace*{-1.5\abovedisplayskip}
\label{fig:preprocessing}
\end{figure}

\begin{figure*}[t]
\includegraphics[width=\textwidth]{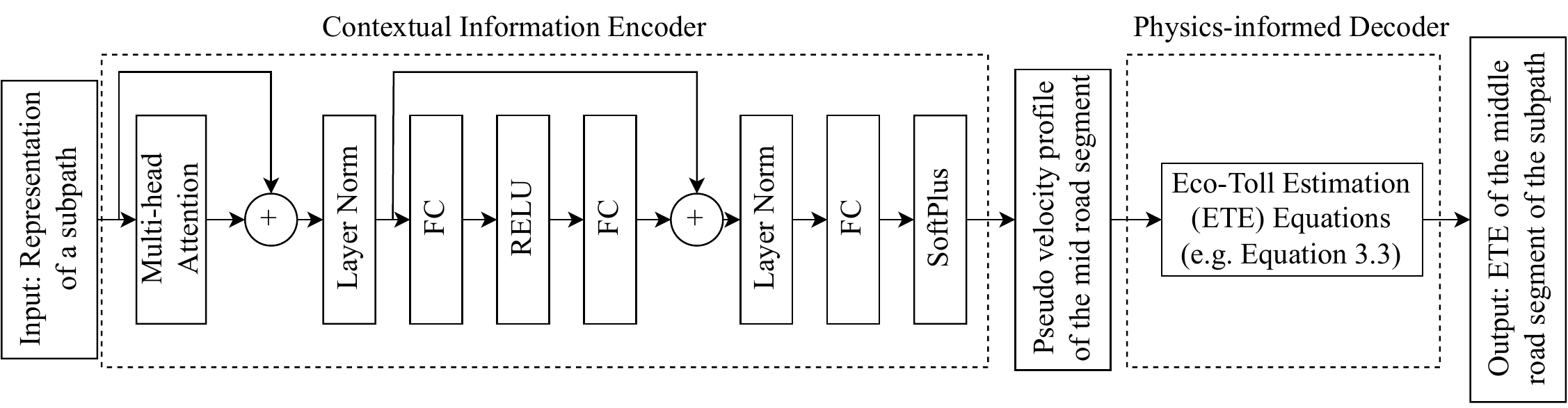}
\vspace*{-2.5\abovedisplayskip}
\caption{Eco-PiNN architecture. Specifically, $\mathrm{softplus}(x)=\mathrm{log}(1+e^x)$. The loss function is detailed in Sec \ref{sec:mtl}.}
\vspace*{-1.5\abovedisplayskip}
\label{fig:network}
\end{figure*}


\paragraph{Road segment spatial proximity feature.} The relative geographic locations of road segments affect the traffic conditions on them, and hence the eco-toll. For example, vehicles on downtown road segments may consume more energy than those on rural road segments because of frequent stops and starts caused by traffic. Thus, we extract this spatial autocorrelation using a road segment spatial proximity extraction module. We first generate an edge-to-vertex dual graph $\mathcal{L(G)}$ (also known as a line graph)   of the original road network $\mathcal{G}$, where a vertex of $ \mathcal{L(G)}$ represents a road segment and an edge of $ \mathcal{L(G)}$ represents a road intersection. Then we use a pre-trained NODE2VEC \cite{grover2016node2vec} model to represent each segment in a $d$-dimensional embedding space (denoted by Figure \ref{fig:preprocessing}). 
After that, the nearby road segments in the road network are given similar representations. 


\paragraph{Categorical features.} We also extract seven categorical features, including five road network attributes: road type,
lane number, bridge or not, and the starting and ending endpoint types;
 as well as two temporal features: departure day and departure time slot. 
Each categorical feature of a road segment is embedded  \cite{gal2016theoretically} into a vector with a pre-defined size (i.e., embedding dimension), and these seven vectors are concatenated together as a $cg$-dimensional representation of the categorical features, where $cg$ represents the sum of the embedding dimensions. All these embedding representations are initialized randomly and learned in the model training stage. 

\paragraph{Numerical features.}Six numerical features are extracted, namely, vehicle mass, speed limit, road length, turning angle to the next road segment in a path, direction angle, and elevation change.
These numerical features are normalized and organized  together as a vector of size $num=6$. By concatenating all the features together, each road segment of the query path can be represented by a vector in $(d+cg+num)$ dimension. 

\paragraph{ETE query representation.} 
To address the challenge that the eco-toll on a segment is influenced by its adjacent segments in the path,  the last preprocessing step represents the query path as a sequence of subpaths to capture the contextual information for each segment using a sliding window. Figure \ref{fig:preprocessing} shows an example that $path_4$ is represented by four subpaths ($subpath_{1-4}$). Specifically, given the contextual window size $w$, the subpath containing the contextual information for the $i$th road segment in a query path is represented by: $[path(i-w),...,path(i),...,path(i+w)]$ (i.e. subpath length $l = 2w+1$). For example, in Figure \ref{fig:preprocessing}, the contextual information for $s_6$ is represented by $subpath_2=[s_9, s_6, s_2]$, given $w=1$. We also implement subpath padding (using zero vectors) to ensure every subpath has the same length. For example, a padding will be added before $s_9$ in $subpath_1$ in Figure \ref{fig:preprocessing}. Finally, we can represent the features of each subpath using a two-dimensional matrix, and each row of the matrix represents the features associated with a road segment, as shown by Figure \ref{fig:preprocessing}. 
\subsection{Eco-PiNN Architecture}

As shown in Figure \ref{fig:network}, the proposed Eco-PiNN architecture is composed of a \textbf{contextual information encoder} to encode the matrix representation of a subpath to a "pseudo" velocity profile of the middle road segment of the subpath, and a \textbf{physics-informed decoder} to decode the "pseudo" velocity profile into the ETE of that road segment. We also propose a novel \textbf{physics-informed jerk penalty regularization} to guide the  training.


\subsubsection{Contextual Information Encoder}

To capture the contextual information in Eco-PiNN, we estimate the eco-toll of a road segment by employing information about its adjacent segments in the given subpath. Attention mechanisms have been widely used to capture interdependence \cite{vaswani2017attention}.
Thus, we design an attention-based encoder to learn the local dependency 
(i.e. how much attention should be given to different road segments in the subpath).
Specifically, the architecture of this encoder is inspired by the encoder module of the Transformer model \cite{vaswani2017attention}, which is composed of a multi-head self-attention mechanism and a fully connected feed-forward network. 
In detail, the input of the contextual information encoder is a subpath represented by $X \in \mathbb{R}^{ (2w+1) \times (d+cg+num)}$ that contains the contextual information for its middle road segment (i.e. the $(w+1)$th segment): $\mathbf{x} = row_{(w+1)}X \in \mathbb{R}^{(d+cg+num)}$.
Then, the feature vector of the segment (i.e. $\mathbf{x}$) is taken as the $query$ of the attention mechanism, and the feature matrix in the corresponding subpath (i.e. $X$) is taken as the packed $keys$ and $values$.
The attention mechanism is formulated as\footnote{In this paper, the head number is set to 1.},
\begin{align}
    Q = \mathbf{x}M^Q, 
    K = XM^K, 
    V = XM^V, \\
    Attention =  (\mathrm{softmax}(\frac{QK^{\mathrm{T}}}{\sqrt{d_k}})V)M^O, \label{eq;attention}
\end{align}
where $M^Q, M^K, M^V, M^O \in \mathbb{R}^{d_k \times d_k}$ are parameter matrices, and $d_k = d+cg+num$ denotes the hidden size of the attention mechanism. Then, the contextual information of the middle road segment can be encoded as $Attention$ by Equation \ref{eq;attention}. 
The encoded contextual information is then fed into a multi-layer perceptron with residual connections and layer-norm to estimate a "pseudo" velocity profile of the middle road segment of the subpath. Note that we use Softplus \cite{glorot2011deep} as the activation function of the last layer of the encoder to avoid zero velocity estimation: $\mathrm{softplus}(x)=\mathrm{log}(1+e^x)$.

\subsubsection{Physics-informed Decoder}\label{piml}

Next, we decode the "pseudo" velocity profile into an eco-toll estimation using a series of eco-toll consumption equations. This physics-informed decoder thus integrates extra human knowledge (e.g. ETE equations) into the neural network, which enables the model to generate more accurate estimation when the training data are limited. 

Equation \eqref{eq:physics_model} shows an example ETE equation which assumes the energy consumption of a vehicle has four parts, namely, the energy used for acceleration, and that needed to overcome the gravitational potential energy change, the rolling resistance at the tires, and the air resistance \cite{cappiello_statistical_2002}.
\begin{equation} \label{eq:physics_model}
W = \frac{m}{\eta} \int (av + gh + c_{rr}gv)dt + \int \frac{A}{2\eta}c_{air}\rho v^3 dt,
\end{equation}
where the energy consumption $W$ is determined by the vehicle's motion properties (i.e., time ($t$), acceleration ($a$), velocity ($v$), and elevation change ($h$)) as well as its physical parameters (i.e., mass ($m$), front surface area ($A$), air resistance coefficient ($c_{air}$), and powertrain system efficiency ($\eta$)). Other symbols in the equation are constants, including gravitational constant $g$, rolling friction coefficient $c_{rr}$, and air density $\rho$.



We begin by defining the "pseudo" velocity profile vector (denoted by $\mathbf{v}$) estimated by the contextual information encoder as a velocity profile on a road segment that is uniformly sampled over time.
Then, under the assumption that the acceleration between velocity samples is uniform (which is reasonable when the length of a velocity profile vector (i.e. $|\mathbf{v}|)$ is large and the travel time between every two velocity samples (denoted by $\Delta t$) is small), we can calculate $\Delta t$ using $\mathbf{v}$ and the length of the road segment $length$ as follows,
\begin{align}\label{eq:parttime}
    \sum_{j=1}^{|\mathbf{v}|-1} \frac{\mathbf{v}(j)+\mathbf{v}(j+1)}{2} *\Delta t = length \nonumber \\
    \Rightarrow \Delta t = 2 * length/\sum_{j=1}^{|\mathbf{v}|-1} (\mathbf{v}(j)+\mathbf{v}(j+1)).
\end{align}
For example, if $\mathbf{v} = [1,2,3,2]$ (m/s) and the length of the road segment is 6.5 meters, then $\sum_{j=1}^{|\mathbf{v}|-1} (\mathbf{v}(j)+\mathbf{v}(j+1))=13$, and $\Delta t  = 1$s.

Then, the acceleration profile can be represented by a vector $\mathbf{a} \in \mathbb{R}^{|\mathbf{v}|}$, and the $j$th acceleration  $\mathbf{a}(j)$ can be calculated as,
\begin{align}\label{eq:acc}
    \mathbf{a}(j) = \frac{\mathbf{v}(j + 1) - \mathbf{v}(j-1)}{2\Delta t}.
\end{align}
Then, the power profile is represented by a vector $\mathbf{p} \in \mathbb{R}^{|\mathbf{v}|}$, where $\mathbf{p}(j)$, the power at the $j$th velocity reading in the velocity profile, can be calculated by
\begin{align}\label{eq:powertrain}
    \mathbf{p}(j) = \frac{m}{\eta} (\mathbf{a}(j)\mathbf{v}(j) + gh + c_{rr}g\mathbf{v}(j)) + \frac{A}{2\eta}c_{air}\rho \mathbf{v}^3(j).
\end{align}
Thus, if $\Delta t$ is small, we can represent the integral in Equation \eqref{eq:physics_model} by the sum of the energy on each time interval which can be calculated by the average power of this time interval times $\Delta t$:
\begin{align}\label{eq:w}
    \hat{W} = \sum_{j=1}^{|\mathbf{v}|-1} \Delta t * \frac{\mathbf{p}(j)+\mathbf{p}(j+1)}{2} .
\end{align}
\textbf{Postprocessing.} The estimated energy consumption of the whole query path can be calculated by summing all the energy estimations of the road segments in the path together.

\subsection{Training Eco-PiNN with Jerk Penalty Regularization and Physics-informed Multitask Learning} \label{sec:mtl}

In the training stage, we introduce a jerk penalty as a regularization to make the estimated "pseudo" velocity profiles more similar to real-world velocity profiles using physics knowledge. We also use a physics-informed multitask learning mechanism to leverages travel time data to guide the training of the Eco-PiNN and to prevent over-fitting. In a multitask
learning mechanism, information is shared across tasks, so the labeled data in all the tasks is aggregated to obtain a
more accurate predictor for each task \cite{zhang2021survey}.

Equation \eqref{eq:loss} is the loss function we use to train the proposed framework. It is a weighted sum of three parts, namely the prediction errors of eco-toll $L_e$, that of travel time $L_t$, and a physics-informed jerk penalty $L_{jerk}$. 
\begin{align}\label{eq:loss}
    L = \omega_e L_e + \omega_t L_t + \omega_{jerk} L_{jerk}.
\end{align}



Inspired by \cite{fang2020constgat}, we define the prediction errors of eco-toll $L_e$ and travel time $L_t$ as a combination of the errors on each road segment and those on the whole path. Specifically, we use the Huber loss \cite{huber1992robust} to represent the error on each road segment, since this loss can help to alleviate the impact of the outliers. Given the predicted and true eco-toll $\hat{W}$ and $W$ on a road segment, the prediction error of the eco-toll on the road segment $L_{seg, e}$ is calculated as follows.
\begin{align}\label{eq:energy loss road segment}
    L_{seg, e} =  \left\{\begin{matrix}
 & \frac{1}{2} (\hat{W}-W)^2 &  |\hat{W}-W| < \delta \\
 & \delta(|\hat{W}-W| - \frac{1}{2}\delta) & otherwise \\
\end{matrix}\right.,
\end{align}
where $\delta$ is a hyperparameter to define prediction outliers. Then we use the mean absolute percentage error (MAPE) to represent the error on the whole path. Given the predicted and true eco-toll $\hat{W}_{path}$ and $W_{path}$ on each path, the prediction error of the eco-toll on a group of paths $L_{path, e}$ is calculated as follows.
\begin{align}\label{eq:energy loss path}
    L_{path, e} =  average(\frac{|\hat{W}^{(k)}_{path} - W^{(k)}_{path}|}{W^{(k)}_{path}}).
\end{align}
Thus, the prediction error of eco-toll $L_e$ is the sum of $L_{seg, e}$ and $L_{path, e}$:
\begin{align}\label{eq:energy loss}
    L_{e} = L_{path, e} +  \frac{1}{n_{path}} \sum_{k=1}^{n_{path}} (\frac{1}{n_{seg}^{(k)}} \sum_{i=1}^{n_{seg}^{(k)}} L_{seg, e}^{(k,i)}),
\end{align}
where $n_{path}$ is the number of paths, $n_{seg}^{(k)}$ is the number of segments in the $k$th path, and $L_{seg, e}^{(k,i)}$ is the prediction error on the $i$th road segment on the $k$th path.


The estimated travel time on a road segment $\hat{t}$ is calculated by $ \hat{t} = (|\mathbf{v}|-1) \cdot \Delta t$. We can get the prediction error of travel time $L_t$ by replacing the predicted and true eco-toll with predicted and true travel time in \Cref{eq:energy loss road segment,eq:energy loss path,eq:energy loss}.
The travel time estimation for a path is the sum of the time estimation of the segments in the path. 

\paragraph{Jerk penalty.} In addition to prediction errors, we introduce a jerk penalty to minimize the jerk of the predicted velocity profiles. Jerk is defined as the first time derivative of acceleration. 
Jerk minimization has been widely used to model driving behavior, with the goal of avoiding high jerk rates that can be uncomfortable to vehicle occupants
\cite{hiraoka_modeling_2005,Singh}. The jerk penalty also serves as a regularization of Eco-PiNN to reduce overfitting. We define the jerk penalty $L_{jerk}$ as the mean of the square of the jerk on each road segment:
\begin{align}\label{eq:jerk penalty}
    L_{jerk} =  \frac{1}{n_{path}} \sum_{k=1}^{n_{path}} \frac{1}{n_{seg}^{(k)}} \sum_{i=1}^{n_{seg}^{(k)}} \sum_{j=1}^{|V|} (\mathbf{jerk}^{(k,i)}(j))^2,
\end{align}
where $\mathbf{jerk}^{(k,i)}(j)$ is the jerk at the $j$th velocity reading in the velocity profile on the $i$th road segment of the $k$th path, and it is calculated as the derivative of the acceleration:
\begin{align}\label{eq:jerk}
    \mathbf{jerk}(j) = \frac{\mathbf{a}(j + 1) - \mathbf{a}(j-1)}{2\Delta t}.
\end{align}

\section{Evaluation}\label{evaluation}
\textbf{Experiment Goals:} We validated Eco-PiNN with (i) a \textit{\textbf{comparative analysis}} to compare the prediction accuracy against several strong baseline methods, (ii)  \textit{\textbf{ablation studies}} to evaluate the contributions of the physics-informed decoder, jerk penalty, contextual information and multitask learning, and (iii) a \textit{\textbf{sensitivity analysis}} to evaluate the impact of 
key parameters (e.g. the weight of jerk penalty). 

\subsection{Experiment Design}

\subsubsection{Data} 
The historical OBD dataset was collected by the Murphy Engine Research Laboratory of the University of Minnesota. It recorded 1343
trips for four diesel trucks in Minnesota operating from Aug. 10th 2020 to Feb. 13th, 2021. The statistical information of these data are detailed in 
Appendix \ref{appendA}.
We divided one day equally into six time slots, and represented the timestamp of entering the road segment by the corresponding time slot.

\begin{figure}[t]
\includegraphics[width=\linewidth]{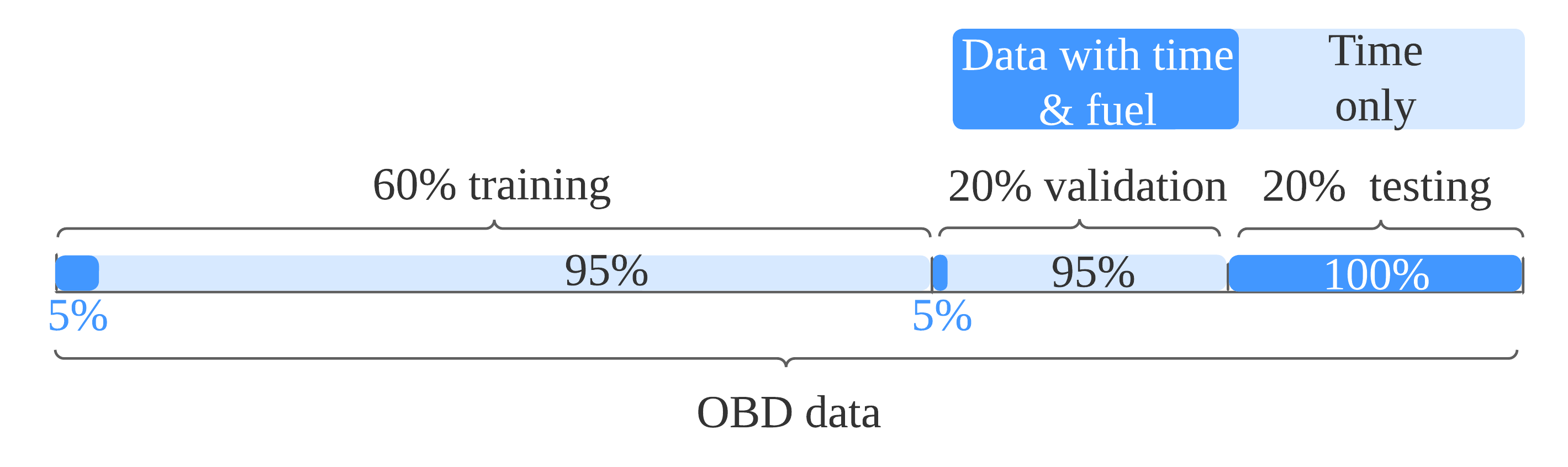}
\vspace*{-2.7\abovedisplayskip}
\caption{Description of how the datasets were split}
\vspace*{-1.5\abovedisplayskip}
\label{fig:dataset}
\end{figure}


 We generated the testing data for our experiments by randomly selecting 20\% of the 1343 vehicle trips. Testing data never changed throughout the experiments, and it contained both travel time and fuel consumption data. To ensure a robust evaluation, the remaining 80\% of the trip data was randomly divided ten different times in ratios of 60\% training and 20\% validation data. Since fuel consumption data is often limited in real world settings, we simulated this challenge by assuming that only a small percentage (e.g., 5\%) of trips (randomly sampled) in any training or validation dataset contained the corresponding ground truth fuel consumption data.
 As noted earlier, the testing data always contained both travel time and fuel information, as shown in Figure \ref{fig:dataset}. 
 Then, we generated datasets containing pairs of eco-toll-estimation (ETE) queries and corresponding travel time and fuel consumption (fuel consumption may have no value). 
 Specifically, each query of all configurations of training/validation datasets corresponded to a sub-trip whose path length was 20, and the step between two sub-trips was set to 5. From the same testing data, different testing datasets were generated based on different settings of the query's path length (from 1 to 200). For each of the ten configurations of our training and validation datasets, we trained the model and tested it using the testing datasets. Finally, we calculated the mean and standard deviation of the estimation error on each testing dataset.


\begin{table*}[]
\centering
\caption{Prediction accuracy when 5\% of training/validation data contained energy consumption.}
\label{tab:comparative5}
\begin{tabular}{@{}lcccccc@{}}
\toprule
\multicolumn{1}{c}{} & \multicolumn{6}{c}{MAPE: Mean (Standard deviation)}                                                                                           \\ \midrule
Path length          & 1                                         & 10    &20                & 50                    & 100                   & 200                   \\ \midrule

NREL                 & 96.45(6.31)          & 28.68(3.07)        & 24.03(2.92) & 20.41(3.30)          & 19.39(3.65)          & 18.83(3.93)          \\
ConSTGAT             & 136.45(8.04)           & 27.51(1.44)       &  23.39(0.90) & 20.55(0.89)         & 19.94(1.74)          & 20.01(2.81)          \\
CI Encoder+FC      &      91.34(6.99) &	25.30(0.86)&	21.95(0.80)	&19.67(0.93)&	19.06(1.31) &	18.81(2.31)     \\
Eco-PiNN                 & \textbf{73.70}(2.37)  & \textbf{21.74}(1.26) &  \textbf{18.50}(1.36) & \textbf{15.83}(1.72) & \textbf{15.13}(1.68) & \textbf{15.78}(1.79) \\ \bottomrule
\end{tabular}
\end{table*}

\begin{table*}[]
\centering
\vspace*{-1\abovedisplayskip}
\caption{Prediction accuracy when 20\% of training/validation data contained energy consumption.}
\label{tab:comparative20}
\begin{tabular}{@{}lcccccc@{}}
\toprule
\multicolumn{1}{c}{} & \multicolumn{6}{c}{MAPE: Mean (Standard deviation)}                                                                                           \\ \midrule
Path length          & 1                                         & 10  & 20                   & 50                    & 100                   & 200                   \\ \midrule

NREL                 & 92.20(3.14)                  & 26.80(0.93)    &22.38(0.89)      & 18.87(0.96)          & 18.35(1.08)          & 18.43(1.56)          \\
ConSTGAT             & 110.01(6.76)               & 23.37(0.55)    & 19.85(0.56)       & 17.47(0.90)          & 17.22(1.21)          & 18.07(1.41)          \\ 
CI Encoder+FC      &      77.27(2.60) 	&	21.18(0.55)&	18.22(0.71)	&15.89(1.14)&	15.20(1.62) &	15.23(1.82)     \\
Eco-PiNN                 & \textbf{70.29(0.89)}  & \textbf{20.56(0.18)} & \textbf{17.34(0.19)} & \textbf{14.68(0.22)} & \textbf{14.12(0.30)} & \textbf{14.86(0.64)} \\ \bottomrule
\end{tabular}
\end{table*}
\vspace*{-1\abovedisplayskip}

\subsubsection{Hyperparameter Settings} 
The embedding size of the NODE2VEC representation of each road segment ($d$) was 32. The walk length was 20. The context size was 10. The number of walks to sample for each node was 10. The p and q parameters in NODE2VEC were set to 1. The number of negative samples used for each positive sample in NODE2VEC was also 1.
The embedding size of the road type and the endpoints type was 4. The embedding size of other categorical features, including starting time, the day of the week, lane number, and bridges, was 2. Thus, the dimension of the aggregated features was 58 (i.e. $d+cg+num=58$). The context window size was $w=1$. 
The output size of the first fully-connected (FC) layer after the attention mechanism was 32. The output size of the second layer was 58, which equaled the dimension of the aggregated features for the residual connection and layer normalization. After that, the output size of the final linear layer  in the encoder was 60 (i.e., $|\mathbf{v}|=60$), and the weights for different loss functions were:  $\omega_e = 0.2$, $\omega_t = 0.8$ and $ \omega_{jerk} = 1e^{-6}$. We used the Adam optimization algorithm \cite{kingma2014adam} to train the parameters with learning rate: $1e^{-4}$ and batch size 512. The parameters were set through a grid search. We used an early stopping mechanism to avoid over-fitting: training was terminated if the model performance stopped improving on the validation set for ten training epochs, after which the best performing model was saved.

\subsubsection{Approaches for Comparison} 
Using mean absolute percentage error (MAPE) as the metric, we compared the prediction accuracy of Eco-PiNN \footnote{Our code: \url{https://github.com/yang-mingzhou/Eco-PiNN} } against three baseline methods

(1) The National Renewable Energy Laboratory (\textbf{NREL}) lookup-table method \cite{holden_trip_2018}. Google Maps claimed that they used the energy estimation models developed by NREL in their recently launched eco-routing function \cite{Google-Maps-Eco-Friendl-Routing}, so we treated this method as a state-of-the-art energy consumption estimation model. It aggregates road segments based on their features and creates a look-up table using the average fuel consumption rate on the aggregated road segments. We used the numerical and categorical features described in Section \ref{sec:preprocessing} to generate the look-up table, and the bin widths for the numerical features were as follows: \textit{mass}: 10000kg; \textit{speed limit}: 10 km/h; \textit{road length}: 100m; \textit{turning angle to the next road segment in a path}: 45 degree; \textit{direction angle}: 45 degree; and \textit{elevation change}: 10 m. 
The fuel consumption rate on unseen road segments was represented by the average fuel rate of its nearest neighbour bin measured by the Euclidean distance.

(2) \textbf{ConSTGAT} \cite{fang2020constgat}. We needed to learn whether the state-of-the-art travel time estimation models will work in an ETE task if physical features are added into the training data. 
 Thus, we implemented ConSTGAT using the same features described in Section \ref{sec:preprocessing}. We treated it as a state-of-the-art travel time eatimation method because it had been deployed in production at Baidu Maps, and successfully served real-world requests \cite{fang2020constgat}. For those parameters that were not mentioned in \cite{fang2020constgat}, we used the similar parameter settings as used by Eco-PiNN, as well as the same early stopping method.

(3) \textbf{CI Encoder+FC}. To verify whether integrating the physics laws with the neural network improves the performance of Eco-PiNN, we developed a model named Contextual Information Encoder + FC (\textbf{CI Encoder+FC}) to conduct an ablation test. This model first encodes the contextual information using the same encoder as Eco-PiNN, and decodes the velocity profile to ETE using a fully-connect layer.

\subsection{Comparative Analysis}
To show how well the methods perform with different amounts of eco-toll information, we tested two settings. In the first setting, 5\% of the queries in the training and validation datasets had corresponding energy consumption data. The results are shown in Table \ref{tab:comparative5}. As can be seen, Eco-PiNN significantly outperformed the baseline methods with all path lengths, especially when the path length was small. For example, when the path length was 1, the Eco-PiNN model was about 20\% more accurate than the state-of-the-art eco-toll estimation model (NREL). When the path length was 200, the Eco-PiNN model was still 3\% more accurate than the baseline methods. 
In the second setting, in Table \ref{tab:comparative20}, the percentage of queries in the training/validation datasets that had corresponding energy consumption data was 20\%.
In this case, given more training data, the accuracy of all methods improved, and Eco-PiNN still outperformed than all baseline methods. 
In conclusion, 
it is reasonable to say that Eco-PiNN significantly outperforms the state-of-the-art methods.

\subsection{Ablation Studies and Sensitivity Analysis}
In this section, we evaluate the contribution of the proposed neural network components on accuracy improvement. In sensitiviy analysis, 5\% of queries in the training and validation datasets had corresponding energy consumption information.

\paragraph{Physics-informed decoder} The contribution of the physics-informed decoder can be analyzed by comparing the performance of Eco-PiNN with that of CI Encoder+FC in both tables. In the first setting shown in Table \ref{tab:comparative5}, Eco-PiNN significantly outperformed CI Encoder+FC (e.g. 18\% more accurate when path length was 1) because of the integration of physics laws.  In the second setting in Table \ref{tab:comparative20}, given more training data, the accuracy of CI Encoder+FC also improved, and Eco-PiNN still outperformed it even though the accuracy difference between them decreased. Nevertheless, the standard deviation of Eco-PiNN under this setting was significantly less than that of CI Encoder+FC, which shows that incorporating the physics laws also improves the stability of the model. In conclusion, 
it is reasonable to say that the integration of physics laws in Eco-PiNN improves the performance and stability.

\begin{figure}[t]
\includegraphics[width=\linewidth]{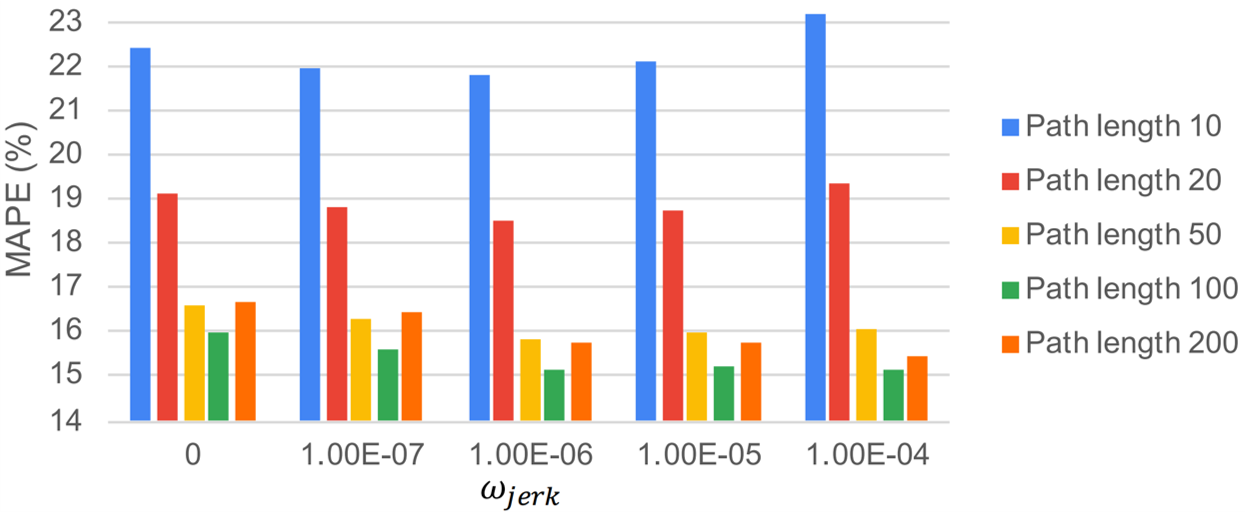}
\vspace*{-2.4\abovedisplayskip}
\caption{Effect of jerk penalty.}
\vspace*{-2.1\abovedisplayskip}
\label{fig:sensitivity_jerk}
\end{figure}

\paragraph{Jerk penalty} To analyze the effect of the proposed jerk penalty, we fixed $\omega_{e}=0.2$ and $w=1$ and varied the weight of the jerk penalty in the loss function $\omega_{jerk}$ from 0 to $10^{-4}$. When $\omega_{jerk} = 0$, the penalty does not affect model training, so the contribution of the penalty can be revealed by comparing the prediction accuracy of the model with $\omega_{jerk} > 0$ and that with $\omega_{jerk} = 0$. The results are shown in Figure \ref{fig:sensitivity_jerk}. The comparison between the MAPE with $\omega_{jerk} = 10^{-6}$ and that with $\omega_{jerk} = 0$ indicates that the jerk penalty component helps improve the estimation accuracy, and the improvement increases with longer paths.

\begin{figure}[t]
\includegraphics[width=\linewidth]{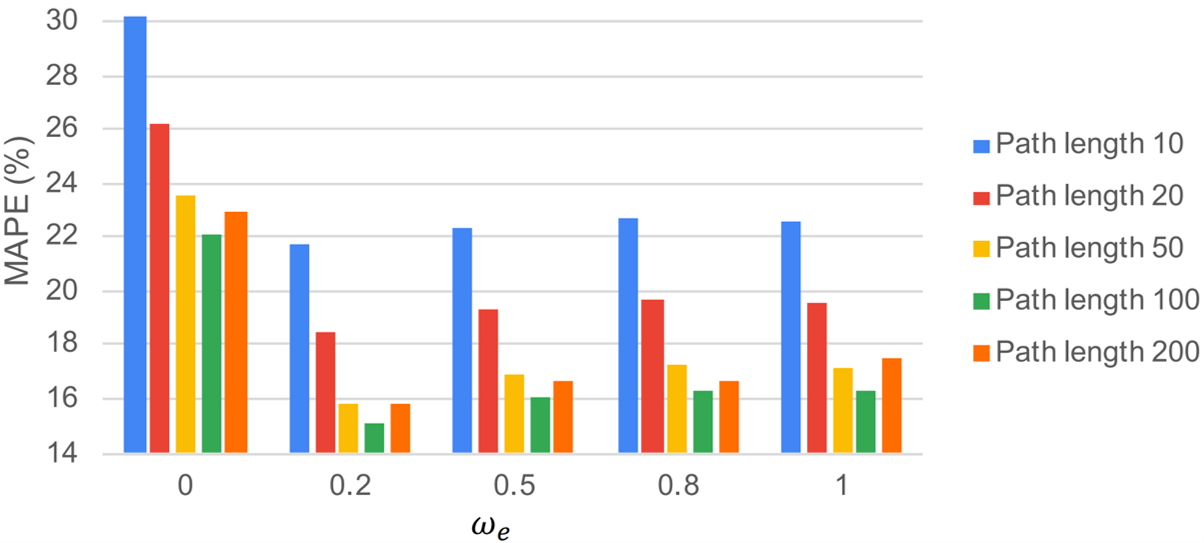}
\vspace*{-2.5\abovedisplayskip}
\caption{Effect of multitask learning component.}
\vspace*{-1.2\abovedisplayskip}
\label{fig:sensitivity_eco_w}
\end{figure}

\paragraph{Multitask learning} 
To analyze the effect of the proposed multitask learning component, we fixed $\omega_{jerk}=1e-6$ and $w=1$ and varied the multitask learning weights (i.e., $\omega_e$ and $\omega_t$), where $\omega_t = 1 - \omega_e$. We varied $\omega_e$ from 0 to 1. When $\omega_e = 1$, the multitask learning component degenerates to an eco-toll estimation task, so the effect of the multitask learning component can be evaluated by comparing the accuracy when $\omega_e < 1$ against that when $\omega_e = 1$. The results are shown in Figure \ref{fig:sensitivity_eco_w}. The comparison between the MAPE with $\omega_e = 0.2$ and that with $\omega_e = 1$ indicates that the multitask learning component helps to improve Eco-PiNN performance, and the improvement increases with increasing path length.

\begin{figure}[t]
\includegraphics[width=\linewidth]{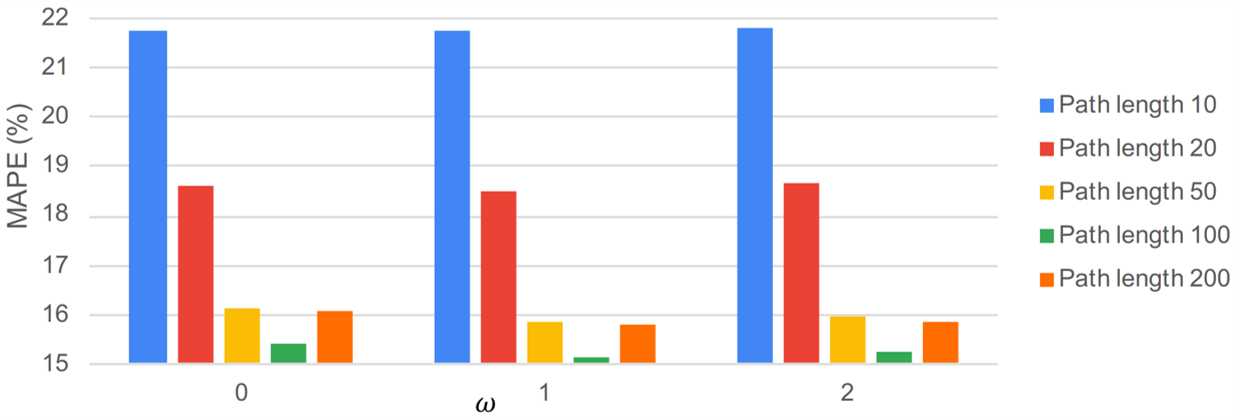}
\vspace*{-2.5\abovedisplayskip}
\caption{Effect of contextual window size.}
\vspace*{-1.8\abovedisplayskip}
\label{fig:sensitivity_window}
\end{figure}

\paragraph{Window size.} We also analyzed the effect of 
the contextual window size by setting $w$ as 0, 1, and 2 and fixed $\omega_{e}=0.2$ and $\omega_{jerk} = 1$. When $w = 0$, no contextual information is considered. The results are shown in Figure \ref{fig:sensitivity_window}. By comparing the MAPE when $w = 0$ with that when $w = 1$, we can see that leveraging the contextual information helped improve the accuracy.

\section{Conclusion}\label{conclusion}
The eco-toll estimation problem quantifies the environmental cost for a vehicle to travel along a path. This problem is of significant importance to society and the environment.
In this work, we propose a novel Eco-toll estimation Physics-informed Neural Network (Eco-PiNN) framework that integrates the physical laws governing vehicle dynamics with a deep neural network. 
Our experiments on real-world vehicle data show that Eco-PiNN yields significantly more accurate eco-toll estimation than state-of-the-art methods. In the future, we plan to 
generate synthetic datasets to analyze the generalization, computational complexity, and sample complexity of Eco-PiNN. We also plan to model the influence of other components (e.g., weather conditions) on eco-toll to further improve its estimation accuracy.

\section*{Acknowledgment}
This material is based upon work supported by the National Science Foundation under Grants No. 1901099, 2147195, and the USDOE Office of Energy Efficiency and Renewable Energy under FOA No. DE-FOA-0002044.  We also thank Kim Koffolt and the Spatial Computing Research Group for valuable comments and refinements. 

\appendix

\section{On-Board Diagnostics (OBD) and Road Network Data}\label{appendA}
Raw OBD data contain a collection of hundreds-of-attributes trajectories and the physical parameters of the corresponding vehicles. Each multi-attribute trajectory records a vehicle's status along a trip and is in the form of a sequence of spatial points, each of which is associated with the vehicle's instantaneous status such as location, exhaust gas emission rate, engine temperature, cumulative energy consumption, and speed. 

The OBD dataset used in this work recorded 1343 trips for four diesel trucks in Minnesota operating from Aug. 10th 2020 to Feb. 13th, 2021 at 1Hz resolution. On average, each trip contained 89.97 road segments. For each road segment, the average length was 608.2157 meters; the average fuel consumption was 0.213421 liter; the average travel time was 28.04122 seconds. In this work, we represented the fuel consumption in the unit of 10 ml (e.g. the average fuel consumption was 21.3421*10ml) so that the prediction errors of the eco-toll task and that of the travel time task were within the same order of magnitude. The mass of the trucks varied with different trips, averaging 23257.71 kg with a standard deviation of 7844.85 kg. Each truck had 10 wheels, with a wheel radius of 0.5003 m. The size of the front area was 10.5 $\mathrm{m}^2$. Engine efficiency was 0.56. The relation between energy consumption and diesel fuel consumption was 38.6MJ (i.e. 10.6KWh) of thermal energy for 1 liter of diesel \cite{davis2021transportation}.

The road network we used was from OpenStreetMap \cite{OpenStreetMap} of Minneapolis with 728548 road segments and 280217 intersections. To calculate the elevation change of each road segment, we used the Esri elevation service \cite{desktop2011release} to capture the elevation of each road intersection in the map (using the mean elevation in a 10m $\times$ 10m cell).

\section{Broadly Related Work}\label{appendB}
\subsection{Eco-toll estimation}
Macroscopic models of eco-toll estimation (e.g., MOVES \cite{kwon_analyzing_2007}) are used to estimate network-wide eco-toll inventories according to aggregated network parameters. By contrast, microscopic models \cite{vijayagopal_automated_2010,brooker_fastsim_2015} estimate a vehicle's instantaneous eco-toll according to physical laws (e.g., classical mechanics and vehicle combustion reaction models) using the vehicle’s velocity profile and physical parameters (e.g., mass and front surface area), as well as some extra information, such as energy source (e.g. diesel or gas). However, because precise velocity profiles are hard to predict due to uncertainty in traffic, microscopic models are mainly used in retrospective research. 

Mesoscopic eco-toll estimation models use the properties of road segments such as average speed and road length as the explanatory variables, and the eco-toll on each road segment as the dependent variable. For example, the National Renewable Energy Laboratory (NREL) proposed a lookup-table-based method, which lists energy consumption rate by category of road segments \cite{holden_trip_2018}. Huang and Peng proposed a Gaussian mixture regression model to predict energy consumption on individual road segments \cite{huang_eco-routing_2018}. Li et al. introduced a physics-guided K-means model that works on paths with historical data \cite{li2018physics}. However, most mesoscopic models are purely data-driven models, which require large amounts of eco-toll data. Also, their results may not be consistent with physical laws and lead to poor generalizability.

\subsection{Travel time estimation}

Our comparative experiments showed that the proposed PiNN model predicts energy consumption more accurately than ConSTGAT, a state-of-the-art travel time estimation (TTE) model. TTE, also known as estimated time of arrival (ETA), aims to estimate a vehicle's travel time for a given path and departure time. Research on the TTE problem mainly focuses on estimating traffic conditions by extracting the spatial-temporal information and the contextual information of a path \cite{wang2018learning,fang2020constgat,derrow2021eta}. TTE models cannot be applied to estimate the environmental cost of a vehicle's travel because: 1) TTE models ignore the physical parameters that affect a vehicle's fuel efficiency, since travel time is mainly affected by traffic conditions. 2) TTE models can be trained by large scale travel time data extracted from Global Positioning System (GPS) trajectory data using the logs of location-based service applications, such as DiDi Chuxing \cite{wang2018learning}, Baidu Maps\cite{fang2020constgat}, and Google Maps\cite{derrow2021eta}. By contrast, the eco-toll data can only be extracted from historical OBD data, or simulated by second-by-second vehicle trajectory data; both of which are limited. The limited availability of such data makes eco toll estimation significantly more challenging than travel time estimation. 
\bibliographystyle{siamplain}
\bibliography{reference}

\end{document}